\documentclass[lettersize,journal]{IEEEtran}
\usepackage[hidelinks, colorlinks=true, linkcolor=blue, citecolor=green]{hyperref}

\usepackage[linesnumbered,ruled,vlined]{algorithm2e}
\usepackage{array}
\usepackage[caption=false,font=normalsize,labelfont=sf,textfont=sf]{subfig}
\usepackage{textcomp}
\usepackage{stfloats}
\usepackage{url}
\usepackage{verbatim}
\usepackage{graphicx}
\usepackage[hidelinks]{hyperref}
\usepackage{cite}
\usepackage{booktabs}
\usepackage{multirow}
\usepackage{textcomp}
\usepackage{amsmath}
\usepackage{makecell}
\usepackage{pifont}
\usepackage{amsmath, amsfonts}

\begin{document}

\title{Label-Free Long-Horizon 3D UAV Trajectory Prediction via Motion-Aligned RGB and Event Cues}

\author{
Hanfang Liang\textsuperscript{1,\dag},
Shenghai Yuan\textsuperscript{2,*\dag}, \textit{Member} IEEE,
Fen Liu\textsuperscript{2},
Yizhuo Yang\textsuperscript{2},\\
Bing Wang\textsuperscript{1},
Zhuyu Huang\textsuperscript{3},
Chenyang Shi\textsuperscript{3},
Jing Jin\textsuperscript{3}
\thanks{\textsuperscript{1}Jianghan University, Wuhan, China.}
\thanks{\textsuperscript{2}Nanyang Technological University, Singapore.}
\thanks{\textsuperscript{3}Beihang University (BUAA), Beijing, China.}
\thanks{\dag\ Equal contribution.}
\thanks{*Corresponding author: Shenghai Yuan (email: shyuan@ntu.edu.sg).}
}




\maketitle

\begin{abstract}
The widespread use of consumer drones has introduced serious challenges for airspace security and public safety. Their high agility and unpredictable motion make drones difficult to track and intercept. While existing methods focus on detecting current positions, many counter-drone strategies rely on forecasting future trajectories, and thus require more than reactive detection to be effective.
To address this critical gap, we propose an unsupervised vision-based method for predicting the three-dimensional trajectories of drones. Our approach first leverages an unsupervised technique to extract drone trajectories from raw LiDAR point clouds, then aligns these trajectories with camera images via motion consistency to generate reliable pseudo-labels. Subsequently, we combine kinematic estimation with a visual Mamba neural network in a self-supervised manner to predict future drone trajectories.
We evaluate our method on the challenging MMAUD dataset, including the V2 sequences featuring wide-FOV, multimodal sensors, and dynamic UAV motion in urban scenes. Extensive experiments show that our framework outperforms supervised image-only and audio-visual baselines in long-horizon trajectory prediction—\textbf{reducing 5-second 3D error by around 40\%}  without using any manual 3D labels. The proposed system provides a cost-effective, scalable alternative for real-time counter-drone deployment, and all code will be released upon acceptance to support reproducible research in the robotics community.

\end{abstract}

\begin{IEEEkeywords}
UAV Trajectory Prediction, Event-Based Vision, Self-Supervised Learning, Long-Horizon Forecasting, Anti-Drone Systems, Multimodal Perception, 3D Pose Estimation, Cross-Modal Alignment, Vision-Mamba, Motion Forecasting
\end{IEEEkeywords}

\section{Introduction}

\begin{figure}[!t]
\centering
\includegraphics[width=3.5in]{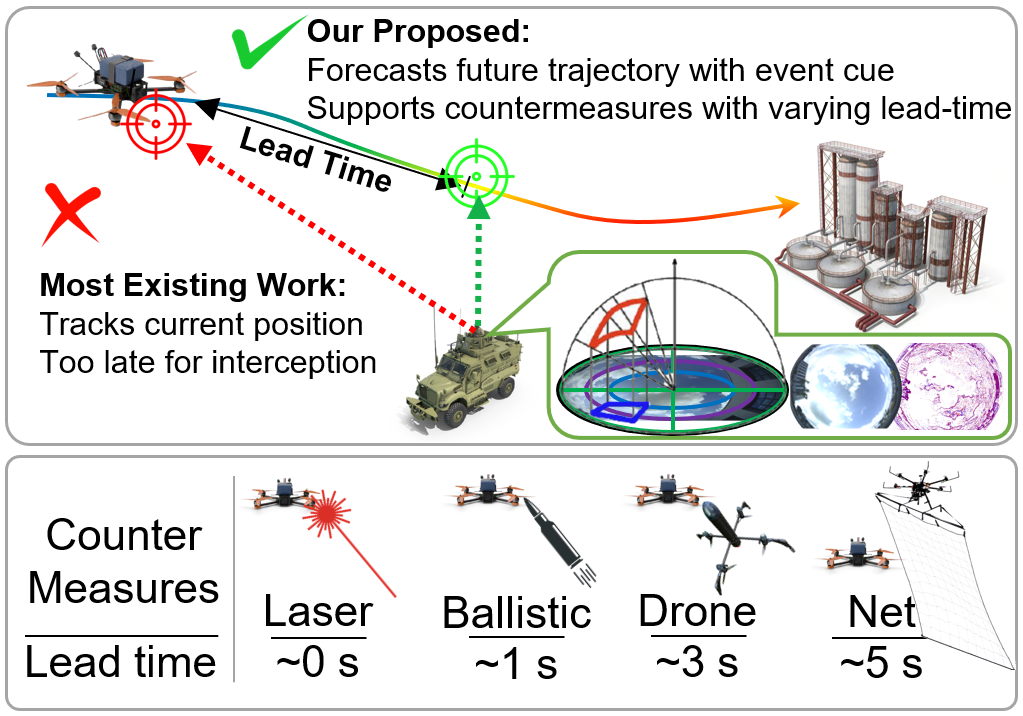}
\caption{
\textbf{Core Motivation: Enabling Lead-Time-Aware UAV Defense.}  
Most existing methods estimate the UAV's current position (red dotted path), offering insufficient lead time to activate physical or autonomous countermeasures. In contrast, our method forecasts the UAV’s future trajectory (green dotted path) based on RGB and event cues, enabling timely response across a range of defense strategies. As illustrated below, different countermeasures—such as lasers \cite{zhao2020design}, ballistic projectiles \cite{cai20247}, drones \cite{liu2025autonomous}, or UAVs equipped with nets \cite{vrba2024onboard}—require varying lead times. This motivates our focus on accurate, long-horizon forecasting under real-world sensing constraints.
}
\label{fig_motivations}
\end{figure}

\IEEEPARstart{T}{he} growing proliferation of unmanned aerial vehicles (UAVs) has driven their adoption in diverse civil domains, including aerial surveillance~\cite{cao2023neptune}, last-mile delivery~\cite{pan2025action}, and infrastructure inspection~\cite{cao2021distributed}. These capabilities are made possible by advances in low-cost platforms~\cite{li2025airswarm}, agile mobility~\cite{ren2025safety}, and autonomous swarm coordination~\cite{zhou2022swarm}. However, the same features that enable legitimate UAV applications have also led to a surge in malicious uses, ranging from privacy concerns~\cite{kurtpinar2024privacy} to airspace disruption~\cite{luan2025fast} and illicit cross-border transport~\cite{krame2023narco}. The wide variability in UAV size, dynamics, and onboard sensing further complicates detection and interception, requiring scalable and adaptive perception systems capable of operating under real-world constraints~\cite{yuan2024mmaud}.

Recent advancements have explored detection and tracking frameworks leveraging radar, LiDAR, audio, and visual modalities~\cite{rahman2024comprehensive, liang2025unsupervised, yang2024av, lei2025audio, da2025new,zhang2025evdetmav}. While these approaches have demonstrated effectiveness in monitoring applications, they predominantly focus on estimating the UAV's current position. However, in proactive defense scenarios, it is crucial to anticipate future trajectories with different lead time—especially for high-speed aerial targets (see Fig. \ref{fig_motivations}). This forecasting task is intrinsically challenging due to the abrupt and nonlinear nature of UAV trajectories~\cite{cao2022direct}, compounded by the scarcity of large-scale, annotated flight datasets~\cite{jiang2021anti}. Although radar and LiDAR offer precise temporal signals~\cite{vrba2024onboard}, their cost and complexity hinder scalability. Meanwhile, vision-based methods are more accessible~\cite{zhao2022vision, zhang2023review}, but remain sensitive to depth ambiguity, dynamic lighting, and the distortions inherent to wide-FOV optics such as fisheye lenses.

Overcoming these \textbf{challenges} requires a new generation of UAV perception systems that move beyond detection toward long-horizon forecasting, while remaining robust to noise, scalable across sensing configurations, and independent of dense manual labeling. In this work, we identify three core challenges: (i) sensor asynchrony across modalities with differing frame rates and timing drift; (ii) strong environmental noise, including glare, motion blur, and multimodal degradation; and (iii) the lack of ground-truth labels for training robust trajectory predictors. 

To address these challenges, we propose an unsupervised, vision-based framework for 3D UAV trajectory prediction. At its core is a Temporal-KNN clustering algorithm that extracts motion-consistent trajectories from raw LiDAR data. These trajectories are then temporally aligned with RGB video frames via a cross-modal motion consistency constraint, enabling the generation of high-quality pseudo-labels without manual annotation or UAV-specific priors.
To support robust learning, we introduce a self-supervised architecture that integrates motion-aware modeling, trajectory projection, and Vision-Mamba-based detection. During training, we incorporate event-based motion cues~\cite{hamann2025etap}, which are simulated from RGB data to approximate the asynchronous perception characteristics of event cameras~\cite{ren2024motion}. This design choice is motivated by two factors: (i) simulated events provide a conservative lower bound, such that any real event sensor is expected to deliver superior motion cues \cite{stoffregen2020reducing}; and (ii) commercial event cameras with both high resolution and wide field of view suitable for UAV-scale tracking were not accessible during development.
At deployment, our system operates solely on video inputs, making it highly scalable and cost-effective for real-time 3D trajectory forecasting in complex aerial environments.

In summary, our main contributions are as follows:

\begin{itemize}
  \item We identify and quantify the impact of sensor asynchrony—a rarely addressed issue in prior works \cite{yang2023av,lei2025audio,yang2024av}—and show that our method achieves robust performance despite the resulting pseudo-label noise.

  \item We present a fully unsupervised pipeline for 3D UAV trajectory extraction and prediction, introducing a Temporal-KNN clustering algorithm that obtains motion-consistent trajectories directly from raw LiDAR data. These trajectories are aligned with video frames through a cross-modal motion consistency constraint, requiring no manual annotations or prior UAV-specific knowledge.

  \item We develop a self-supervised multimodal learning architecture that integrates motion-aware modeling, trajectory projection alignment, and Vision-Mamba-based detection. To emulate asynchronous supervision, we incorporate event motion cues.

  \item We evaluate our framework on a real-world dataset and demonstrate that it outperforms state-of-the-art supervised and multimodal baselines across various sensing conditions, including low light and challenging dynamic scenes.
\end{itemize}

\section{Related Works}

\subsection{UAV Detection and Estimation}
UAV detection and pose estimation have been explored across various sensing modalities, including LiDAR, visual, and multimodal systems. Several LiDAR-centric methods have demonstrated the utility of spatio-temporal consistency in point cloud sequences. For example, Deng et al. [29] accumulated sequential LiDAR frames to identify dynamic UAV points through clustering, subsequently applying LSTM networks for trajectory-aware center regression, with further refinement via Kalman filtering. In contrast, Zheng et al. [30] addressed UAV pose estimation using only visual input, predicting predefined image-space keypoints followed by visual feature extraction for classification and 6-DoF tracking. Extending to thermal modalities, Lan et al. [31] proposed an RGB-T tracking framework based on unsupervised learning, illustrating the effectiveness of sparse thermal fusion in enhancing UAV localization. The application of event cameras in UAV detection remains relatively underexplored but has recently emerged as a promising research direction~\cite{tofighi2025survey, rossi2025event, da2025new}. However, existing studies are largely constrained to short-range scenarios, typically within 10 meters, limiting their applicability to real-world aerial surveillance tasks.

Recent advancements in Transformer architectures have further improved UAV tracking performance, particularly in challenging scenarios involving small-scale or agile targets. UTTracker, introduced by Yu et al. [32], integrates local tracking, global detection, and background suppression modules within a unified Transformer-based framework. This design significantly improves performance under cluttered and dynamic backgrounds. Similarly, Li et al. [33] presented Aba-ViTrack, a one-stream Vision Transformer (ViT) model that fuses template-search coupling with adaptive token filtering, achieving both computational efficiency and tracking robustness.

Beyond traditional sensing, audio-visual multimodal networks have emerged as promising alternatives for UAV trajectory estimation. Vora et al. [28] leveraged a Teacher-Student paradigm, using visual networks to generate pseudo-ground truth for training audio-based regressors. Ding et al. [34] further proposed a two-stage framework, where audio cues were first used for coarse localization, followed by visual refinement to achieve improved accuracy. These studies highlight the benefits of cross-modal supervision and complementary sensor fusion for enhanced estimation.

\begin{figure}[!t]
\centering
\includegraphics[width=3in]{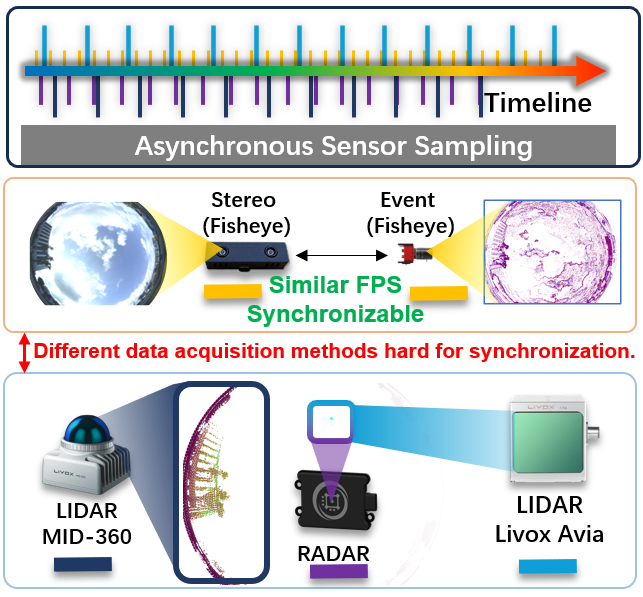}
\caption{ Illustration of the heterogeneous sensing setup and the lack of temporal synchronization between sensors. Due to unsynchronized timestamps among LiDARs (Avia, MID360), RADAR, Event and fisheye cameras, cross-modal alignment suffers from temporal inconsistency, introducing significant pseudo-label noise in trajectory estimation.}
\label{fig_1}
\end{figure}

\subsection{Trajectory Forecasting for Aerial Targets}
The task of trajectory forecasting, particularly for UAVs, is of growing interest due to its relevance in autonomous navigation, airspace management, and anticipatory tracking. A variety of sequence modeling approaches have been proposed to address the inherently multimodal and nonlinear dynamics of UAV motion.

Becker et al. [35] trained RNN-based predictors on synthetic UAV trajectories generated from quadrotor dynamics and optimal control, demonstrating generalizability to real-world aerial paths. To better capture spatial heterogeneity, Zhou et al. [36] introduced a spatial-temporal learning aggregator (STLA) with degenerate self-attention, which outperformed conventional MLP and Transformer-based architectures in short-term forecasting tasks.

To address multimodal uncertainty in future motion prediction, Makansi et al. [37] proposed a reachability-prior framework guided by semantic maps and egomotion planning. This approach enables more plausible and semantically consistent future state predictions. Zhang et al. [38] explored the use of ADS-B data for UAV trajectory modeling via LSTM networks, achieving accurate mid-term forecasts in large-scale air traffic scenarios. More recently, Zhong et al. [39] developed a 4D trajectory prediction method that performs spatio-temporal clustering and applies CNN-LSTM pipelines for segment-specific prediction. Their approach demonstrates superior accuracy in 0–3s forecasting horizons when compared with LSTM, GRU, and velocity extrapolation baselines.

\section{Proposed Methods}
In our proposed method, the pipeline is divided into two main components: unsupervised sparse point cloud extraction and multimodal detection fusion. First, we employ a clustering-based evaluation approach to extract UAV trajectories from LiDAR data while simultaneously assessing points from different LiDAR sensors within the point cloud. Based on the extracted trajectories, we perform a dynamic analysis to predict the UAV's motion, generating pseudo ground-truth trajectories that adhere to the UAV's kinematic constraints.

Subsequently, we project the LiDAR data onto images and integrate it with RGB images and event camera data. The event data is computed by capturing pixel changes between consecutive image frames. The pseudo ground-truth labels obtained from unsupervised point cloud extraction are used as supervision signals for training. Furthermore, we propose a multimodal tracking and prediction architecture based on the Mamba framework, which fuses RGB, event, and LiDAR modalities to accurately track and predict UAV motion.

\subsection{Problem Definition}

Let \(P\) denotes a set of points sampled at discrete timestamps \(t\), \(P=\left\{p_{i}^{t}|i=1,2,\cdots,N_{t},t=1,2,\cdots,T\right\}\). Let \(P_{i}^{t}\) denote the 3D coordinates of the \(i\)-th point in the \(t\)-th frame. Let \(N_{t}\) represent the total number of points in the point cloud at frame \(t\).The total number of frames is denoted by \(T\).

Let the state vector \(\mathcal{X}_{3}\) encapsulate the UAV's 3D kinematic states in discrete time, represented as:
\begin{equation}
    \mathcal{X}_{3}=[t, \mathcal{V}, \alpha]^{T}, \mathcal{X} \in \mathbb{R}^9
\end{equation}
where t denotes the position, \(\mathcal{V}\) denotes the velocity, and \(\alpha\) denotes the acceleration in the world coordinate frame.

To relate the 3D state to the 2D image space, we adopt a unified projection model suitable for fisheye cameras. The 3D point \(t=[x,y,z]^{T}\) is projected to pixel coordinates  \([u,v]^{T}\) via the following steps:
\begin{align}
x' &= \frac{x}{z + \xi \sqrt{x^2 + y^2 + z^2}}, \\
y' &= \frac{y}{z + \xi \sqrt{x^2 + y^2 + z^2}}, \\
r^2 &= x'^2 + y'^2, \quad \delta_r(r) = 1 + k_1 r^2 + k_2 r^4, \\
\tilde{x} &= x' \cdot \delta_r(r), \quad \tilde{y}= y' \cdot \delta_r(r), \\
u &= f_x \cdot \tilde{x} + c_x, \quad v = f_y \cdot \tilde{y} + c_y
\end{align}

Where \(k_1\) and \(k_2\) are the first and second-order radial distortion coefficients, and \(\xi\) denotes the parameter of the unified imaging model \cite{ahmad20143d}, which accounts for the offset of the projection center along the optical axis, \(\delta_r(r)\) is the radial distortion function.

Let \(\pi_{3\rightarrow2}(\cdot)\) denotes the projection operator, which maps a 3D position to image coordinates under fisheye distortion and intrinsic transformation.

To compute image-space motion, we differentiate the projection function with respect to time using the chain rule. The components \(\dot{\mathbf{u}}\) and \(\ddot{\mathbf{u}}\) represent the predicted pixel velocity and acceleration from 3D dynamics.
  The image-plane velocity is computed as:

\begin{equation}
  \dot{\mathbf{u}} = \frac{d\pi_{3\rightarrow2}(\mathbf{t})}{d\mathbf{t}} \cdot \mathcal{V} = \mathbf{J}_{\mathbf{t}} \cdot \mathcal{V}  
\end{equation}

Where \(\mathbf{J}_{\mathbf{t}} \in \mathbb{R}^{2\times3}\) is the Jacobian of the projection with respect to the 3D position. Similarly, the image acceleration is given by:
\begin{equation}
\ddot{\mathbf{u}} = \mathbf{J}_{\mathbf{t}} \cdot \boldsymbol{\alpha} + \dot{\mathbf{J}}_{\mathbf{t}} \cdot \mathcal{V}
\end{equation}

Hence, the complete 2D projected motion state denotes as \(\mathcal{X}_{2}\).
\begin{equation}
\mathcal{X}_{2} = 
\begin{bmatrix}
\mathbf{u} \\
\dot{\mathbf{u}} \\
\ddot{\mathbf{u}}
\end{bmatrix}
=
\begin{bmatrix}
\pi_{3\rightarrow2}(\mathbf{t}) \\
\mathbf{J}_{\mathbf{t}} \cdot \mathcal{V} \\
\mathbf{J}_{\mathbf{t}} \cdot \boldsymbol{\alpha}
\end{bmatrix}
\in \mathbb{R}^6.
\end{equation}

\begin{figure}[!t]
\centering
\includegraphics[width=3in]{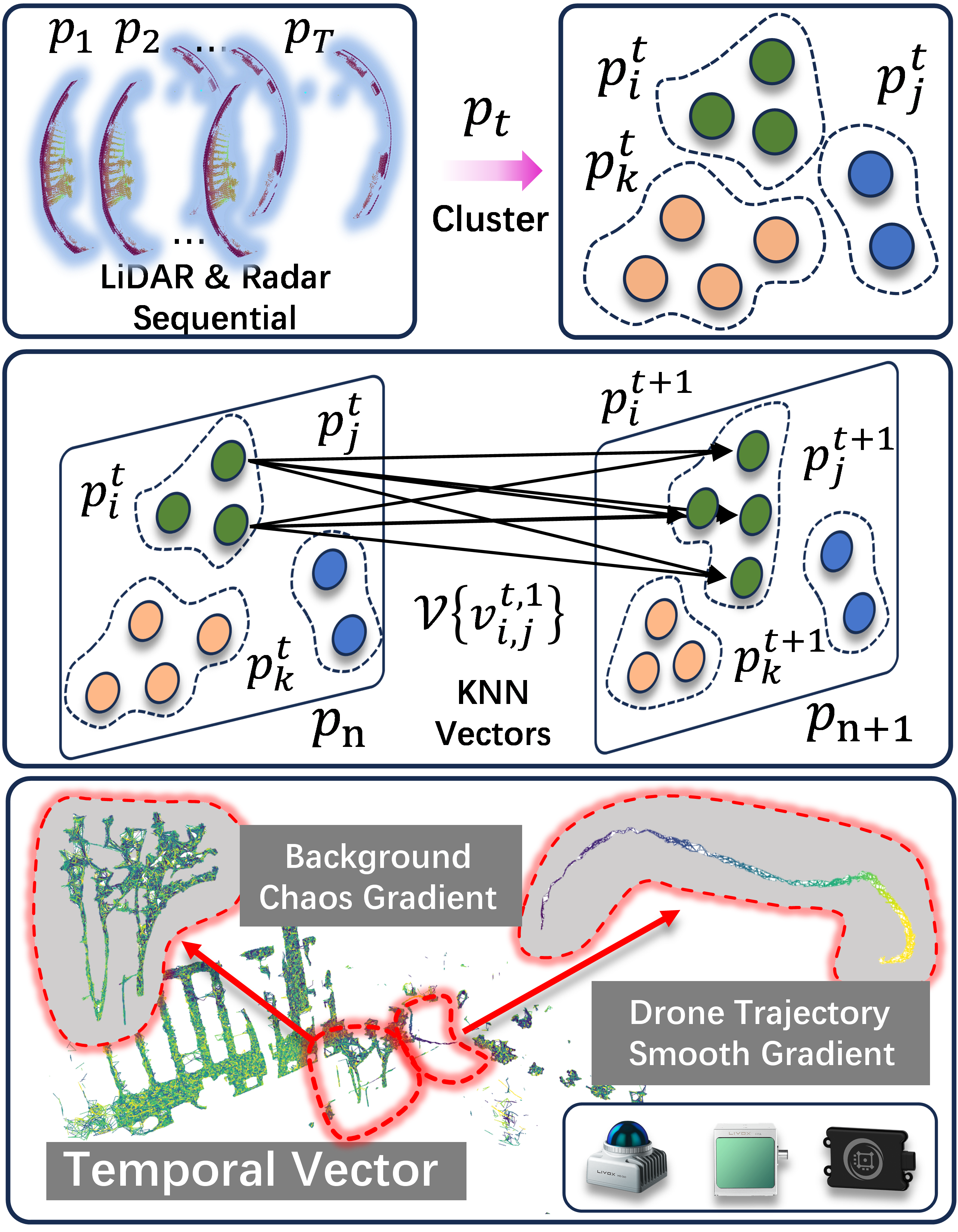}
\caption{Unsupervised temporal filtering via KNN vector gradient consistency. Sequential LiDAR and radar frames are clustered, and temporal KNN vectors are computed between point clusters. Background and static objects exhibit inconsistent or chaotic motion patterns, while the UAV trajectory forms a smooth and coherent motion gradient. This enables filtering of noisy points and isolation of plausible motion trajectories without annotation.}
\label{fig_2}
\end{figure}

\subsection{Temporal KNN Trajectory Extraction}


\begin{figure*}[!t]
\centering
\includegraphics[width=1\textwidth]{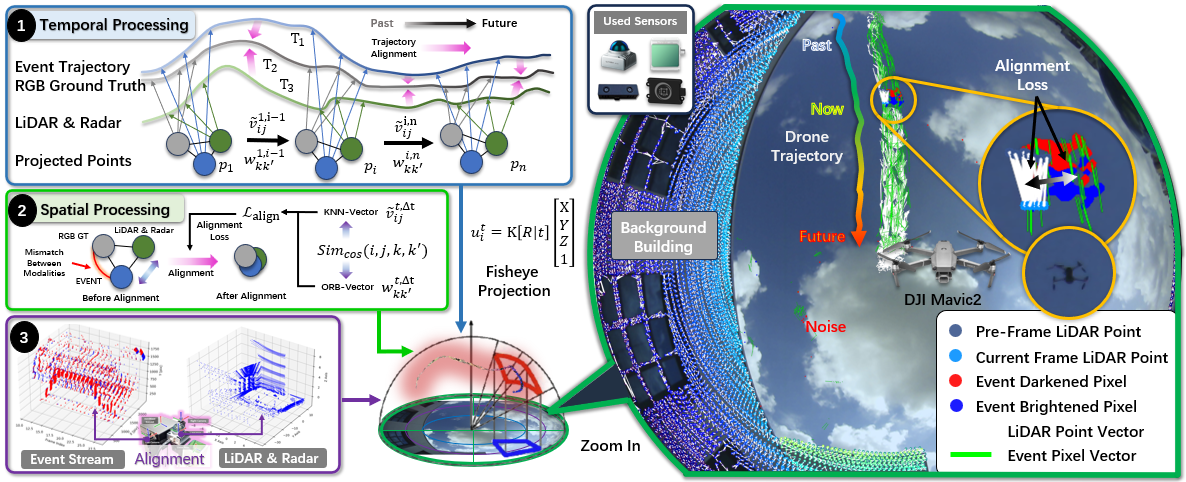}
\caption{Illustration of cross-modal motion alignment for pseudo-label refinement. Raw LiDAR and radar trajectories are projected into the fisheye image frame, but exhibit pixel-level misalignment with visual motion cues due to calibration and temporal drift. We align 3D temporal vectors with ORB or event-based image-space vectors using cosine and full-state loss, as shown in the zoomed region. This significantly reduces projection error and improves the reliability of 2D motion supervision.}
\label{fig_3}
\end{figure*}

\begin{algorithm}[t]
\caption{Temporal Vector KNN Construction with Gradient Filtering}
\KwIn{Point clouds $\{P^t\}_{t=1}^{T}$; parameters $K$, $\Delta t$, gradient threshold $\tau$}
\KwOut{Filtered temporal vector set $V$}

\For{$t = 1$ \KwTo $T - \Delta t - 1$}{
  \ForEach{$p_i^t \in P^t$}{
    Find $\text{KNN}_1(p_i^t) \subseteq P^{t + \Delta t}$\;
    Find $\text{KNN}_2(p_i^t) \subseteq P^{t + \Delta t + 1}$\;
    Initialize $G_i \leftarrow 0$\;

    \ForEach{paired point $(p_j^{t+\Delta t}, p_j^{t+\Delta t+1})$}{
      $\vec{v}_{ij}^{(1)} \leftarrow p_j^{t+\Delta t} - p_i^t$\;
      $\vec{v}_{ij}^{(2)} \leftarrow p_j^{t+\Delta t+1} - p_i^t$\;
      $\nabla \vec{v}_{ij} \leftarrow (\vec{v}_{ij}^{(2)} - \vec{v}_{ij}^{(1)}) / \Delta t$\;
      $G_i \leftarrow G_i + \left\| \nabla \vec{v}_{ij} \right\|$\;
    }

    Compute $g_i \leftarrow G_i / K$\;

    \If{$g_i \leq \tau$}{
      \ForEach{$p_j^{t + \Delta t} \in \text{KNN}_1(p_i^t)$}{
        Add $v_{ij}^{t, \Delta t} = p_j^{t + \Delta t} - p_i^t$ to $V$\;
      }
    }
  }
}
\Return $V$\;
\end{algorithm}

The raw drone trajectories were acquired using multiple sensors, including Livox Avia, Mid360 LiDAR, and millimeter-wave radar. To fully utilize and extract information from these sensors, we aimed to fuse data from Livox Avia, Mid360, and radar. However, direct fusion is challenging due to discrepancies in timestamps across these sensors, as illustrated in Fig. \ref{fig_1}.

A simple timestamp alignment approach would inadvertently amplify the temporal gaps between sensors, resulting in estimation errors for the drone’s state during subsequent kinematic analysis. Additionally, projecting point clouds onto images becomes increasingly erroneous due to these temporal mismatches. The cumulative effect of these errors significantly destabilizes the pseudo labels derived through unsupervised methods.

To address these issues, we propose a novel temporal vector clustering method inspired by K-Nearest Neighbors (KNN), as shown in Fig. \ref{fig_2}. While traditional KNN methods have been widely used in conventional point cloud detection and segmentation networks due to their consistent dimensionality inputs, our approach emphasizes temporal motion rather than static point cloud frames. Specifically, we connect point clouds across consecutive frames using a temporal KNN approach, recording the direction and magnitude of vectors linking these frames. 

Although sensor timestamps may appear chaotic, objects in motion maintain temporal consistency. Leveraging this physical property, our method effectively mitigates the errors arising from temporal mismatches across different sensors. Moreover, for moving objects, vectors connecting successive frames exhibit a high consistency with the object's direction of motion, whereas stationary objects, such as buildings and trees, generate vectors that appear random and unordered. Thus, we utilize this vector-based strategy to filter the point clouds effectively.

Specifically, we define a spatial distance function between two points as \(d(p_i^t, p_j^{t + \Delta t})\). 

\begin{equation}
    d(p_i^t, p_j^{t + \Delta t}) = \| p_i^t - p_j^{t + \Delta t} \|_2
\end{equation}

At each timestamp \(t\), we construct a temporal k-nearest neighbors (KNN) set by finding the \(K\) closest points as \(\text{KNN}\left ( \cdot \right ) \) in the subsequent frame at \(t+\delta t\) for each point \(p_{i}^{t}\), denotes as \(\mathcal{K}\).

\begin{equation}
\mathcal{K}(p_i^t, K, \Delta t) = \left\{ p_j^{t + \Delta t} \;\middle|\; \text{KNN} \left ( d(p_i^t, p_j^{t + \Delta t}) \right ) \right\}
\end{equation}

Based on the above definition, we construct the set of temporal vectors as \(\mathcal{V}\).

\begin{equation}
\mathcal{V}=\left\{ v_{ij}^{t, \Delta t} \;\middle|\; v_{ij}^{t, \Delta t}=p_j^{t +\Delta t}-p_i^t,\;p_j^{t+\Delta t}\in \mathcal{K}(p_i^t,K,\Delta t)\right\}
\end{equation}


\begin{algorithm}[t]
\caption{Directional and Full-State Consistency Between Projected KNN Vectors and ORB Motion}
\label{alg:knn_orb_alignment}
\KwIn{%
3D point cloud frames $\{P^t\}_{t=1}^T$, ORB keypoints $\{O^t\}_{t=1}^T$,\\
projection model $\mathbf{P}(\cdot)$, time offset $\Delta t$, neighbors $K$%
}%
\KwOut{Cosine similarity $\text{Sim}_{\text{cos}}$, full-state alignment loss $\mathcal{L}_{\text{align}}$}

\For{$t = 1$ \KwTo $T - 2\Delta t$}{
    \textbf{Step 1: Extract projected motion vectors} $p_i^t$\;
    \ForEach{$p_i^t \in P^t$}{
         $\mathcal{K} \leftarrow \mathcal{K}(p_i^t, K, \Delta t)$\;
        \ForEach{$p_j^{t+\Delta t} \in \mathcal{K}$}{
            $u_i^t \leftarrow \mathbf{P}(p_i^t)$,\quad $u_j^{t+\Delta t} \leftarrow \mathbf{P}(p_j^{t+\Delta t})$\;
            $\tilde{v}_{ij}^{t,\Delta t} \leftarrow u_j^{t+\Delta t} - u_i^t$\;
            Estimate $\dot{u}_{ij}$, $\ddot{u}_{ij}$ by finite difference\;
            $\mathcal{X}_2^{(i)} \leftarrow [u_i^t, \dot{u}_{ij}, \ddot{u}_{ij}]$\;
        }
    }

    \textbf{Step 2: Extract ORB motion vectors}\;
    $\mathcal{M}^{t,\Delta t} \leftarrow \{(o_k^t, o_{k'}^{t+\Delta t})\}$\;
    \ForEach{$(o_k^t, o_{k'}^{t+\Delta t}) \in \mathcal{M}^{t,\Delta t}$}{
        $w_{kk'} \leftarrow o_{k'}^{t+\Delta t} - o_k^t$\;
        Estimate $\dot{o}_{kk'}$, $\ddot{o}_{kk'}$\;
        $\mathcal{X}_2^{\text{ORB},(k)} \leftarrow [o_k^t, \dot{o}_{kk'}, \ddot{o}_{kk'}]$\;
    }

    \textbf{Step 3: Align and compute similarity} $\tilde{v}_{ij}^{t,\Delta t}$\;
    Find nearest ORB keypoint $o_k^t$ to $u_i^t$\;

    \[
    \text{Sim}_{\text{cos}} = 
    \frac{\tilde{v}_{ij}}{\|\tilde{v}_{ij}\|_2} \cdot 
    \frac{w_{kk'}}{\|w_{kk'}\|_2}
    \]

    Compute state alignment loss:
    \[
    \mathcal{L}_{\text{state}} = \left\| \mathcal{X}_2^{(i)} - \mathcal{X}_2^{\text{ORB},(k)} \right\|_2^2
    \]

    \textbf{Step 4: Aggregate}\;
    $\mathcal{L}_{\text{align}} \leftarrow \sum \left( 1 - \text{Sim}_{\text{cos}} + \lambda \cdot \mathcal{L}_{\text{state}} \right)$\;
}
\end{algorithm}

Using temporal KNN, we capture the short-term motion between consecutive frames by linking spatially proximate points across time, forming a local temporal representation of moving points.

Given a point \( p_i^t \in P^t \), we find its $K$ nearest neighbors in the future frames \( P^{t+\Delta t} \) and \( P^{t+\Delta t + 1} \). 
\begin{equation}
\vec{v}_{ij}^{(1)} = p_j^{t+\Delta t} - p_i^t, \quad 
\vec{v}_{ij}^{(2)} = p_j^{t+\Delta t+1} - p_i^t
\end{equation}

The temporal gradient of the motion vector is defined as \(\nabla \vec{v}\).
\begin{equation}
\nabla \vec{v}_{ij} = \frac{\vec{v}_{ij}^{(2)} - \vec{v}_{ij}^{(1)}}{\Delta t}
\end{equation}

We compute the mean gradient magnitude \(g_i\) for each point.
\begin{equation}
g_i = \frac{1}{K} \sum_{j=1}^K \left\| \nabla \vec{v}_{ij} \right\| = 
\frac{1}{K \cdot \Delta t} \sum_{j=1}^K \left\| \vec{v}_{ij}^{(2)} - \vec{v}_{ij}^{(1)} \right\|
\end{equation}

A vector \( \vec{v}_{ij}^{(1)} \) is preserved only if \(g_i \leq \tau\). where \( \tau \) is a threshold controlling the temporal smoothness of motion.

Using this unsupervised approach, we extract drone trajectories with motion vectors, those vectors embedded motion information directly from LiDAR and Radar data.

\subsection{Unsupervised Trajectory Projection Alignment}

After extracting the drone trajectories, we further project LiDAR data onto images using a series of projection matrices \(\mathbf{P}\).



Most existing methods project point clouds directly onto images. Despite we using calibration projection matrices \(\mathbf{P}\) to map 3D point clouds into 2D pixel coordinates, misalignment occurs due to inaccuracies in calibration matrices and inherent sensor errors, particularly when projecting LiDAR and radar data. 

Such misalignment can lead to catastrophic results when employing unsupervised methods, as initial trajectory extraction errors are significantly magnified during projection. To avoid these issues, instead of directly projecting the raw point clouds, we first project the trajectory vectors obtained in the previous section.

Although point cloud data and event camera data differ considerably in their modalities, our unsupervised method inherently captures spatial and motion characteristics analogous to those captured by event cameras, which detect pixel-level intensity changes between successive image frames. Therefore, we project the extracted trajectory vectors \(\mathcal{V}\) onto the event cues image data.

To project motion vectors, we take the temporal KNN vector:

\begin{equation}
v_{ij}^{t, \Delta t} = p_j^{t + \Delta t} - p_i^t
\end{equation}

Directly projecting a 3D vector using the projection matrix can lead to geometric distortion due to the non-linearity introduced by the camera model, especially perspective projection. Therefore, instead of projecting the vector itself, we first project the endpoints \(p_j^{t + \Delta t}\) and \(p_i^t\) into the image space to obtain their corresponding pixel locations \(u_j^{t + \Delta t}\) and \(u_i^t\).We then project both endpoints into image space and define the 2D motion vector as \(\tilde{v}_{ij}\).

\begin{figure}[!t]
\centering
\includegraphics[width=3.4in]{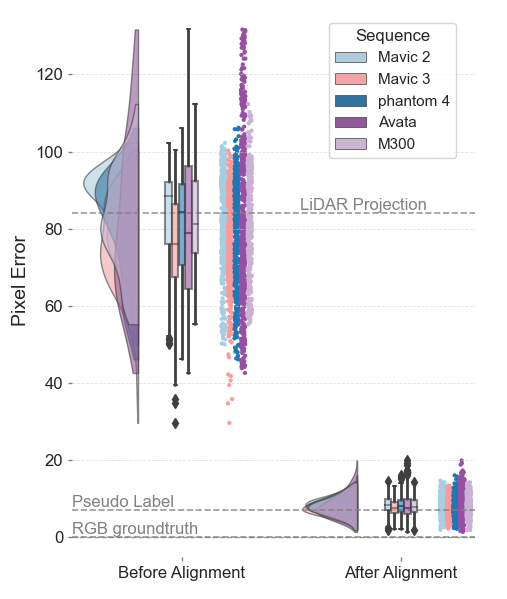}
\caption{Comparison of direct 3D projection versus motion-aligned supervision. Directly projecting LiDAR trajectories onto the image yields large errors due to sensor noise and miscalibration. By aligning with image-space ORB motion, we filter noisy trajectories and obtain significantly improved pixel-level consistency for supervision.}
\label{fig_4}
\end{figure}

Additionally, similar processing is required for the event camera data. Analogous to point cloud vectors, we calculate pixel-level motion vectors in 2D images by analyzing changes in pixel intensities. We use the ORB algorithm to detect keypoints and compute pixel-level vectors. The projected LiDAR vectors and the event cue vectors demonstrate geometric similarity in their 2D vector features, as shown in Fig. \ref{fig_3}.

To evaluate the consistency between 3D kinematic predictions and observed 2D visual motion, we incorporate dense ORB-based keypoint tracking as reference motion supervision.

We define the set of ORB keypoints extracted from the image at time \(t\) as \(O^{t}\).

\begin{equation}
    O^t = \left\{ o_k^t \;\middle|\; k = 1, \ldots, M_t \right\}, \quad o_k^t = 
\begin{bmatrix}
u_k^t \\
v_k^t
\end{bmatrix}^T
\in \mathbb{R}^2
\end{equation}

To estimate image motion based on visual features, we extract ORB keypoints and pairwise matches between frames \(t\) and \(t+\Delta t\) denotes as \(\mathcal{M}^{t, \Delta t}\).
\begin{equation}
\mathcal{M}^{t, \Delta t} = \left\{ \left( o_k^t, o_{k'}^{t + \Delta t} \right) \;\middle|\; o_k^t \leftrightarrow o_{k'}^{t + \Delta t} \right\}
\end{equation}

Then, for each pair of matched keypoints, the 2D ORB flow vector is computed as \(w_{k k'}^{t, \Delta t}\).
\begin{equation}
    w_{k k'}^{t, \Delta t} = o_{k'}^{t + \Delta t} - o_k^t
\end{equation}

Although both sets of 2D motion vectors—projected from 3D point clouds and extracted from ORB feature matches—capture the local motion between frames, they exhibit a degree of noise and irregularity.

To establish cross-modal consistency between the predicted motion from 3D temporal structure and the observed motion from image-space features, We wish to align each predicted \(\mathcal{X}_{2}\) with the best-matching observation vector by comparing not only motion direction, but also position and acceleration consistency. we align the full 2D projected motion state \(\mathcal{X}_{2}\) computed from both 3D point trajectories and ORB keypoint tracks.


The image space motion state of the ORB tracks is defined as\(\mathcal{X}_2^{\text{ORB}}\).
\begin{equation}
\mathcal{X}_2^{\text{ORB}} =
\begin{bmatrix}
o_k^t \\
w_{kk'}^{t,\Delta t} \\
\left(o_{k''}^{t+2\Delta t} - 2 o_{k'}^{t+\Delta t} + o_k^t\right) / \Delta t^2
\end{bmatrix}
\in \mathbb{R}^6
\end{equation}

The motion alignment loss is then formulated as a minimum full-state distance as \(\mathcal{L}\).
\begin{equation}
\mathcal{L} = \sum_{i} \min_{j} \left\| \mathcal{X}_2^{(i)} - \mathcal{X}_2^{\text{ORB},(j)} \right\|_2^2
\end{equation}

This formulation enforces consistency between the predicted and observed image motion states, capturing trajectory dynamics across multiple time steps and modalities.

Leveraging motion consistency across modalities, we align and calibrate these vectors, significantly reducing feature misalignment inherent to unsupervised methods.

As shown in Fig. \ref{fig_4}, we compare projections from the original LiDAR trajectory points, ground-truth LiDAR points, ground-truth image bounding boxes, and our matched points obtained through LiDAR-image alignment. This strategy substantially improves the accuracy of projected points derived using unsupervised methods.

By converting inter-modal motion consistency into pixel-level vector consistency, we generate aligned unsupervised label data suitable for subsequent image-based training.

\subsection{Self-Supervised Learning Frame}


\begin{figure*}[!t]
\centering
\includegraphics[width=1\textwidth]{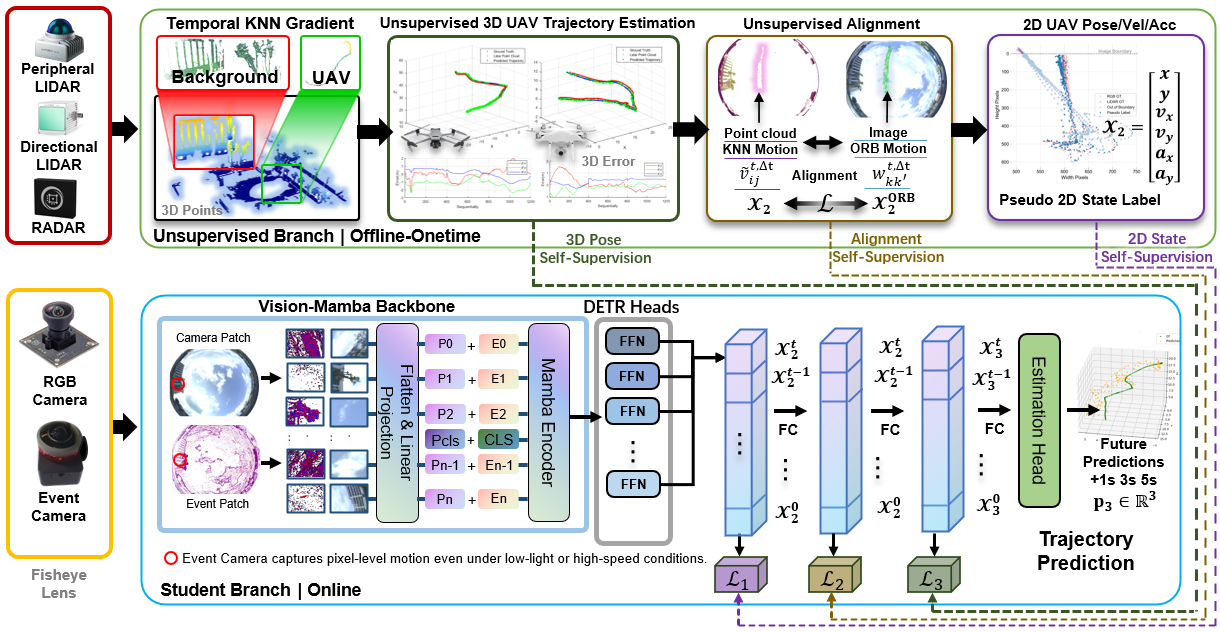}
\caption{Overview of the proposed label-free long-horizon self-supervised UAV trajectory prediction framework. The offline unsupervised branch extracts 3D UAV trajectories from multimodal sensors via Temporal-KNN gradient filtering, followed by cross-modal alignment with ORB-based visual motion to generate pseudo 2D pose labels. The online student branch uses a Vision-Mamba backbone to fuse RGB and event inputs for trajectory prediction, supervised by alignment-consistent pseudo labels across multiple stages. Event cues enhance pixel-level motion perception under low light and fast dynamics.}
\label{prediction2}
\end{figure*}

Most existing methods are limited to 2D image-based object detection or monocular 6-DoF pose estimation, which are insufficient for practical UAV detection scenarios. This issue is particularly pronounced when using wide-angle or fisheye cameras, where severe distortion occurs near image boundaries. As a result, the perceived position and size of a UAV in the image can be significantly affected by lens distortion, which directly compromises the accuracy of depth estimation. Consequently, UAVs of different physical sizes may appear within the same image region and exhibit similar bounding box dimensions, yet reside at vastly different altitudes in the real world.

Estimating the 3D pose of UAVs in real-world environments typically relies on auxiliary depth sensors, which are substantially more costly and bulky compared to standard cameras. For example, a typical fisheye camera costs around \textbf{\$100}, while commercial LiDAR units such as the \textit{Mid-360} and \textit{Livox Avia} are priced at approximately \textbf{\$769} and \textbf{\$2,542}\footnote{Price obtained from Google search before publications.}, respectively. This cost disparity makes LiDAR-based solutions impractical for large-scale or mobile deployment scenarios.


To overcome this limitation, we propose a multimodal Mamba-based prediction architecture within a hierarchical self-supervised framework. This model integrates RGB and event modalities to perform UAV detection and 3D trajectory prediction without requiring depth supervision. As illustrated in Fig.~\ref{prediction2}, the proposed framework progressively learns structural representations at multiple levels, mitigating error accumulation typically caused by noisy pseudo-labels in unsupervised settings.

Due to the unavailability of commercially accessible event cameras with both high resolution and wide field of view during development, we adopt a simulated event generation strategy by extracting pixel-level intensity changes from RGB inputs. While approximate, these simulated cues provide effective motion edges for supervision, as confirmed by our ablation studies. Crucially, this design offers a conservative lower bound on performance: real event cameras are known to produce significantly cleaner and more temporally precise motion responses~\cite{stoffregen2020reducing}, suggesting that our framework is likely to benefit further from real event input without modification. This positions our system as future-proof and ready for enhanced deployment once such hardware becomes readily available.

Specifically, our pipeline employs stage-wise self-supervised training using 2D pseudo-labels, geometric alignment transformations, and projected 3D supervision. This design enables the model to extract accurate semantic representations from the unsupervised data obtained in Sections C and D. Given the Vision-Mamba's output—which includes UAV class, scale, and image-space location—we first apply a transformation to align it with the projected LiDAR points, and then infer the 3D position of the UAV by estimating its real-world scale and depth based on its 2D footprint.

In the first stage, we perform self-supervised training of the Vision-Mamba backbone using aligned 2D pseudo-labels derived from RGB images.The input fisheye images \(\mathcal{I}_{c}\) and event camera frames \(\mathcal{I}_{e}\) are jointly processed by the Vision-Mamba backbone to extract high-level multimodal visual features.  

RGB images and event cues are passed through separate branches of the network, with their features fused via cross-attention layers to enhance spatiotemporal representation.

To effectively encode the complementary modalities of RGB and event data, we adopt a dual-branch patch embedding module. The RGB image \(\mathcal{I}_{img} \in \mathbb{R}^{H\times W \times C}\) and the event image \(\mathcal{I}_{evt} \in \mathbb{R}^{H\times W \times C}\) are each processed into spatial patches and projected into a common feature dimension.

We denote the resulting patchified tokens as \(\mathcal{I}_{img,p},\mathcal{I}_{evt,p}  \in \mathbb{R}^{J\times (P^{2} \cdot C)}\), where \(P\) is the patch size and \(J= \frac{HW}{P^{2}}\) is the total number of patches.
\begin{equation}
\mathbb{I}_{img}^{0} = [\mathcal{P}_{img}^{cls};\mathcal{I}_{img,p}\Lambda_{img,p}] + E_{pos}^{img}
\end{equation}
\begin{equation}
\mathbb{I}_{evt}^{0} = [\mathcal{P}_{evt}^{cls};\mathcal{I}_{evt,p}\Lambda_{evt,p}] + E_{pos}^{evt}
\end{equation}

where \(\Lambda_{img,p}\in \mathbb{R}^{(p^{2} \cdot C ) \times D} \) and \(\Lambda_{evt,p}\in \mathbb{R}^{(p^{2} ) \times D} \) are learnable linear projection weights; \(\mathcal{P}_{\cdot}^{cls} \in \mathbb{R}^{D}\) denotes the learnable class token for each modality; \(E_{pos}^{\cdot} \in \mathbb{R}^{(J+1)\times D}\) are modality-specific positional embeddings.

Each modality-specific embedding \(\mathbb{I}_{\cdot}^{0}\) is then passed through a Vision-Mamba encoder independently, where \(L\) is the number of Mamba layers.
\begin{equation}
\mathbb{I}_{img}^{L}=Mamba_{img}(\mathbb{I}_{img}^{0}),\quad\mathbb{I}_{evt}^{L}=Mamba_{evt}(\mathbb{I}_{evt}^{0})
\end{equation}

After obtaining separate representations from the image and event branches, we fuse the modalities using a cross-attention mechanism, followed by a lightweight fusion block to jointly reason across streams.

Given encoded representations from each stream \(\mathbb{I}_{img}^{L}\in \mathbb{R}^{J\times D}\) and \(\mathbb{I}_{evt}^{L}\in \mathbb{R}^{J\times D}\)
we define a cross-attention mechanism \(\mathcal{A}\)that uses image tokens as queries and event tokens as keys and values.

\begin{equation}
\mathcal{A} = \text{Softmax} \left( \frac{ \left( \mathbb{I}_{\text{img}}^L W_Q \right) \left( \mathbb{I}_{\text{evt}}^L W_K \right)^{\top} }{ \sqrt{d} } \right)
\end{equation}
\begin{equation}
    \mathbb{F} = \mathcal{A} \cdot \left( \mathbb{I}_{\text{evt}}^L W_V \right)
\end{equation}

And where \(W_Q, W_K, W_V \in \mathbb{R}^{D\times D'}\) is the attention weight matrix; \(\mathbb{F} \in \mathbb{R}^{J\times D'}\) represents the fused output from cross-attention. The final fused token is obtained by concatenating or adding the two branches and applying a Fusion Block, where the Fusion Block is a residual MLP block.
\begin{equation}
\text{FusionBlock}(x) = \text{MLP}\left( \text{LayerNorm}(x) \right) + x
\end{equation}

After the fusion of dual-stream representations from RGB and event modalities, the fused tokens are passed into a Transformer decoder, following the Detection Transformer (DETR) framework.

We initialize a fixed number of learnable object queries:
\begin{equation}
    \mathcal{Q} = \left\{ \mathbf{q}_m \in \mathbb{R}^{D'} \;\middle|\; m = 1, \ldots, M \right\}, \quad \mathcal{Q} \in \mathbb{R}^{M \times D'}
\end{equation}

These object queries attend to the fused token embeddings via cross-attention in each decoder layer:
\begin{equation}
    \mathbf{Z}^{(l)} = \text{TransformerDecoder}^{(l)}\left( \mathbf{Z}^{(l-1)}, \mathbb{F}_{\text{final}} \right), \quad \mathbf{Z}^{(0)} = \mathcal{Q}
\end{equation}

Where \(\mathbf{Z}^{(0)}=\mathcal{Q}\)  and the decoder is stacked for \(L\) layers. The final decoder output \(\mathbf{Z}^{(L)}\in \mathbb{R}^{M\times D'}\) is passed into two parallel feed-forward heads for classification \(\mathbf{p}_m \) and bounding box regression \(\hat{\mathbf{b}}_m \), where \(\mathbf{p}_m \in \mathbb{R}^{C+1}\) represents class logits, \(\hat{\mathbf{b}}_m \in [0,1]^{4}\)  is the normalized bounding box in \((x,y,w,h)\) format.
\begin{equation}
    \mathbf{p}_m = \text{MLP}_{\text{cls}}(\mathbf{z}_m), \quad \hat{\mathbf{b}}_m = \text{MLP}_{\text{bbox}}(\mathbf{z}_m)
\end{equation}

During training, bipartite matching is applied between predicted and ground truth objects, and the total loss \(\mathcal{L}_{\text{DETR}} \) is a sum of classification and L1 and GIoU regression loss.
\begin{equation}
    \mathcal{L}_{\text{DETR}} = \mathcal{L}_{\text{cls}} + \lambda_1 \mathcal{L}_{\text{L1}} + \lambda_2 \mathcal{L}_{\text{GIoU}}
\end{equation}
Several weighting factors in our framework govern the balance between different loss components. In the DETR-style detection loss, the coefficients $\lambda_1$ and $\lambda_2$ control the relative importance of the regression terms: $\mathcal{L}_{\text{L1}}$ encourages precise bounding box localization, while $\mathcal{L}_{\text{GIoU}}$ enforces shape and spatial overlap consistency. These weights allow the network to jointly optimize for spatial precision and geometric robustness. Separately, the alignment loss employs a hyperparameter $\lambda$ to balance directional similarity (via cosine loss) against full-state dynamic consistency (position, velocity, and acceleration) during cross-modal supervision between projected LiDAR motion and ORB-based image cues. All three coefficients are selected empirically based on validation performance, and we observe that careful tuning of these values improves convergence stability and final trajectory accuracy. In future work, these hyperparameters could be further optimized using reinforcement learning \cite{messikommer2024reinforcement} or gradient-based hyperparameter optimization \cite{maclaurin2015gradient}.

Then in the second stage, we design a feedforward neural network \(f^{(2)}:\mathbb{R}^{2} \rightarrow \mathbb{R}^{2}\) to map the center \(\mathbf{c}=[x_{center},y_{center}]\) of the detected bounding box \(\hat{\mathbf{b}}_m\) to its corresponding image projection coordinates \(\mathbf{p}=[x_{proj},y_{proj}]\).To facilitate this mapping, we first buffer the output token sequence \(\mathbb{I}_n \in \mathbb{R}^{J\times D}\) buffer by the Vision-Mamba backbone for all consecutive frames. These sequences are then temporally aggregated to recover the motion state of the detected object across frames. Specifically, we compute the 2D kinematic state vector. In parallel, we estimate the corresponding projected kinematic state. This stage is trained with pseudo-labels from \(\mathcal{X}_2\) and \(\mathcal{X}_2^{ORB}\).As a result, the network learns to approximate projection behavior while preserving consistency with the observed 2D temporal dynamics.

The network consists of two hidden layers and ReLU activation, and linear layer denotes as \(\mathbf{L}\), ReLu activation layer denotes as \(\mathbf{R}\).
\begin{equation}
    f^{(2)}(\mathbf{p},\tilde{\mathcal{X}_2})=\mathbf{L}_{6}(\mathbf{R}(\mathbf{L}_{128}(\mathbf{R}(\mathbf{L} _{6}(c,\tilde{\mathcal{X}_2^{ORB}})))))
\end{equation}

During this stage, only \(f^{(2)}\) is trained, and the mean square error (MSE) loss is used.

The third stage uses another feedforward network \(f^{(3)}:\mathbb{R}^{2} \rightarrow \mathbb{R}^{3}\), which takes the image-projected coordinates \(\mathbf{p}\) and 2D trajectory state \(\mathcal{X}_2\) from stage 2; width and height of bounding box \(\left \{ \hat{\mathbf{b}}_m|w,h \right \}\) and UAV class \(\mathbf{p}_m\) from stage 1 as inputs and predicts the corresponding 3D position \(\mathbf{x}=[x,y,z]\) and 3D state \(\mathcal{X}_3\).

\begin{equation}
f_3(\tilde{\mathbf{x}},\tilde{\mathcal{X}_3}) = \mathbf{L}_{9}(\mathbf{R}(\mathbf{L}_{64}(\mathbf{R}(\mathbf{L}_{128}(\mathbf{p},\tilde{\mathcal{X}_2},\mathbf{p}_m,w,h)))))
\end{equation}

This stage recovers the 3D trajectory based on fused 2D inputs and is supervised by pseudo-ground-truth 3D positions from point cloud registration.

\subsection{Motion Estimation Head}
After the 3D position of the target UAV is regressed from the center of the detected bounding box and the associated features, we introduce a simple motion estimation head to predict its future trajectory.

This module is implemented as a small feedforward neural network that takes the estimated current state and timestamps as input, and predicts the UAV's position at several future timestamps. Specifically, the input includes the current 3D position \(\mathbf{x}_t = [x_t, y_t, z_t]\), UAV 2D and 3D states \(\mathcal{X}_2, \mathcal{X}_2^{ORB}, \mathcal{X}_3\) , the corresponding time \(t\), and a normalized time delta \(\Delta t\) indicating the prediction horizon. And outputs predicted positions \(\hat{\mathbf{x}}_{t+\Delta t}\) for \(\Delta t = 1s, 2s, 3s, 5s\). It is trained using standard Mean Squared Error (MSE) loss between the predicted and pseudo-ground-truth 3D positions.

\begin{figure*}[!t]
\centering
\includegraphics[width=1\textwidth]{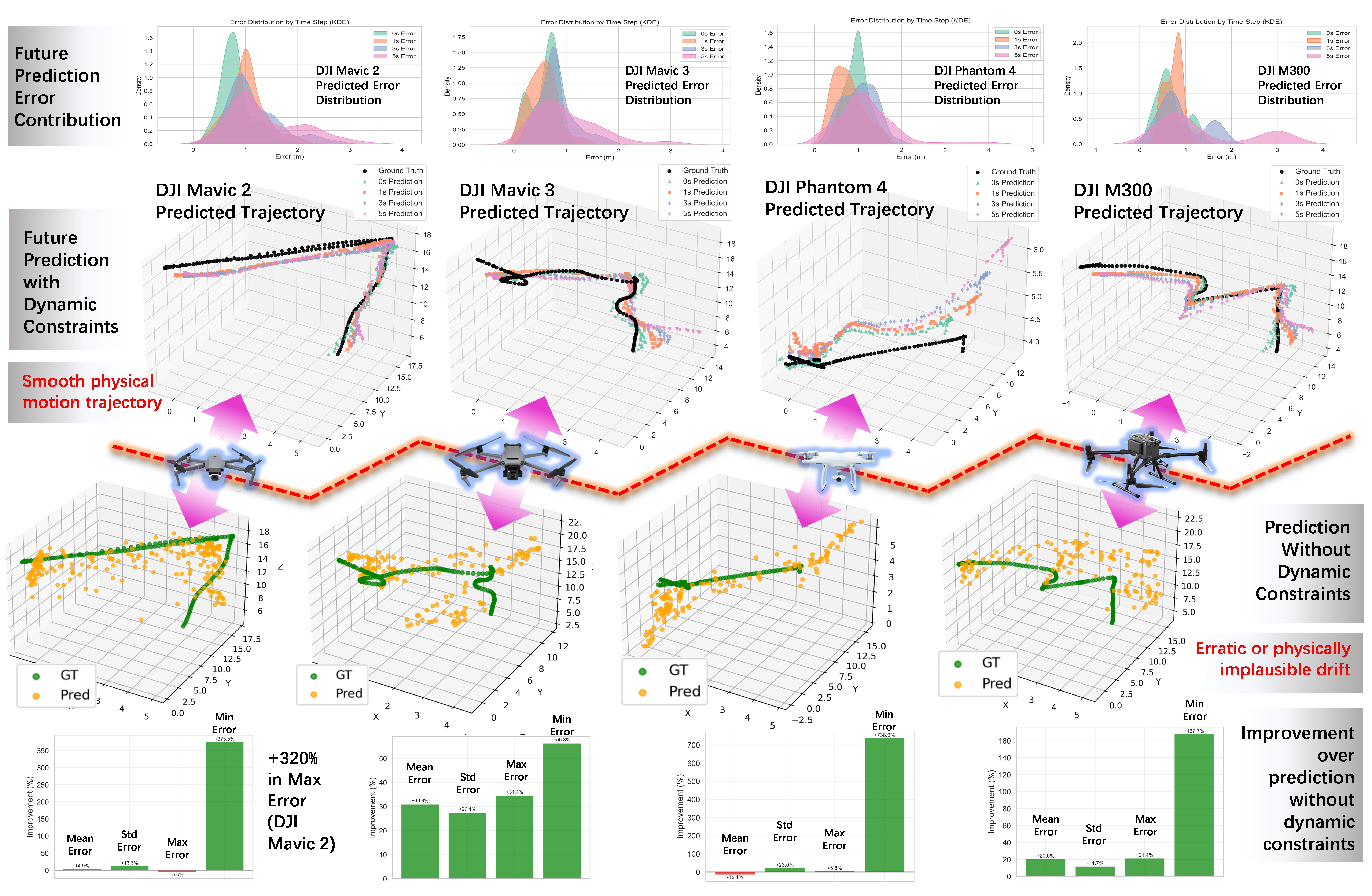}
\caption{Comparison of future trajectory prediction under dynamic constraints across four drone platforms (Part types illustration of Datasets). 
Top row shows the predicted error distribution at future time steps. 
Middle two rows compare predicted 3D trajectories with and without dynamic constraints.
Bottom row illustrates the relative improvement in prediction error.
This figure illustrates a subset of UAV categories included in the dataset.}
\label{Results}
\end{figure*}

\begin{table}[t]
\centering
\caption{Error comparison (in meters) and relative change from manual, no unaligned and aligned labels.}
\label{tab:align_improvement}
\renewcommand{\arraystretch}{1.1}
\begin{tabular}{lcccc}
\toprule
\textbf{Metric} & Manual & No Align. & Pseudo (Ours) & $\Delta_{\text{No Align↑}}$ \\
\midrule
$D_x$ & 0.32 & 1.28 & 0.43  & +30.4\textperthousand \\
$D_y$ & 0.48 & 5.64 & 0.96  & +58.2\textperthousand \\
$D_z$ & 0.28 & 3.27 & 0.51  & +64.6\textperthousand \\
$E_{0s}$ & 0.64 & 6.72 & 1.19  & +57.2\textperthousand \\
$E_{1s}$ & 0.98 & 11.32 & 2.53  & +44.7\textperthousand \\
$E_{3s}$ & 3.94 & 18.54 & 5.41  & +34.2\textperthousand \\
$E_{5s}$ & 7.62 & 80.96 & 10.40  & +77.8\textperthousand \\
\bottomrule
\end{tabular}
\vspace{4pt}

\footnotesize{
These results are evaluated on the MMAUD 2D dataset, the pseudo-label alignment strategy demonstrates consistent improvements over unaligned labels.
}
\end{table}

\section{Experiments}

\begin{figure}[!t]
\centering
\includegraphics[width=\linewidth]{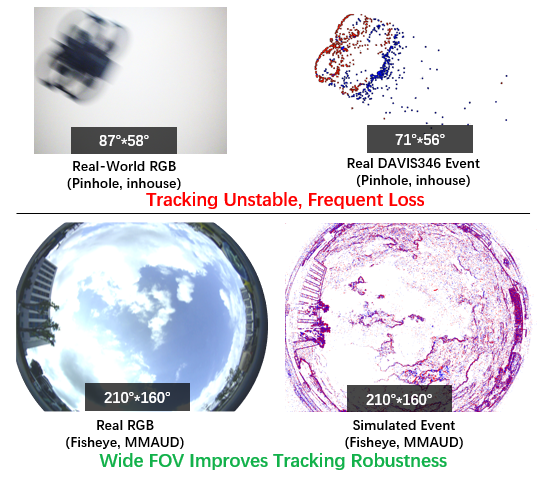}
\caption{From limited FOV to robust tracking: a comparison of sensing configurations.
Top: Our initial in-house tests using a pinhole RGB camera and DAVIS346 event sensor reveal the limitations of narrow field-of-view (87°×58°, 71°×56°), leading to unstable and frequently lost UAV tracking.
Bottom: The MMAUD dataset addresses these limitations by providing wide-FOV (210°×160°) real fisheye RGB and simulated event data, significantly enhancing tracking robustness.}
\label{simulated_data}
\end{figure}

\subsection{Evaluation Dataset}
We initially collected a small-scale real-world dataset using a pinhole RGB camera and a DAVIS346 event-based sensor (Fig.~\ref{simulated_data}). However, the narrow field of view (87\textdegree$\times$58\textdegree{} for RGB and 71\textdegree$\times$56\textdegree{} for DAVIS346) and limited resolution made it unsuitable for wide-area UAV tracking, often resulting in unstable detection and frequent target loss. While newer event cameras—such as the Prophesee GenX720—now offer high resolution (1280$\times$720) and ultra-wide FOVs exceeding 180\textdegree, such hardware was not yet available during our development cycle.

To overcome these limitations, we adopted the publicly available \textbf{MMAUD dataset}~\cite{yuan2024mmaud}, which provides a rich set of multimodal sensors and significantly broader scene coverage. MMAUD includes high-frame-rate fisheye RGB video at 1280$\times$720 with a 210\textdegree$\times$160\textdegree{} field of view—closely mirroring the characteristics of modern wide-FOV event cameras. Although it does not include real event data, we approximate asynchronous motion cues using simulated event streams derived from the RGB modality. This strategy offers a conservative training setup that ensures compatibility with real event sensors, which are expected to further improve downstream performance due to their cleaner and more temporally accurate motion signals~\cite{stoffregen2020reducing}.

To the best of our knowledge, \textbf{MMAUD} is the only publicly available anti-UAV dataset that combines \emph{multiple LiDARs, radar, stereo fisheye RGB, simulated event cues, synchronized audio, and Leica MS60-based ground truth} across diverse urban sites and UAV platforms. This makes it a uniquely suitable benchmark for real-world UAV perception and long-horizon trajectory forecasting under noisy, dynamic, and multimodal conditions.

In this work, we utilize both the V1 and V2 sequences from MMAUD. The \textbf{V1 subset} features UAV flights within a 30-meter radius and has been widely adopted in prior literature \cite{lei2025audio, varghese2024yolov8, wang2024yolov10, yang2024av, yang2023av, vora2023dronechase, wang2022visualnet, tao2021someone, bochkovskiy2020yolov4}. The \textbf{V2 subset}, in contrast, includes larger-scale UAV trajectories extending up to 100 meters. Although V2 has been less frequently used in published works, it was featured in the CVPR Anti-UAV Challenge\footnote{\href{https://cvpr2024ug2challenge.github.io/track5.html}{ CVPR Challenge:  \url{https://cvpr2024ug2challenge.github.io/track5.html}}}, where the top-performing solution achieved an average 3D localization error of approximately \textbf{2 meters}. Despite its increased difficulty, V2 serves as a valuable testbed for evaluating long-horizon generalization and trajectory forecasting in more expansive, unconstrained scenarios.

\begin{figure*}[t]
\centering
\includegraphics[width=7in]{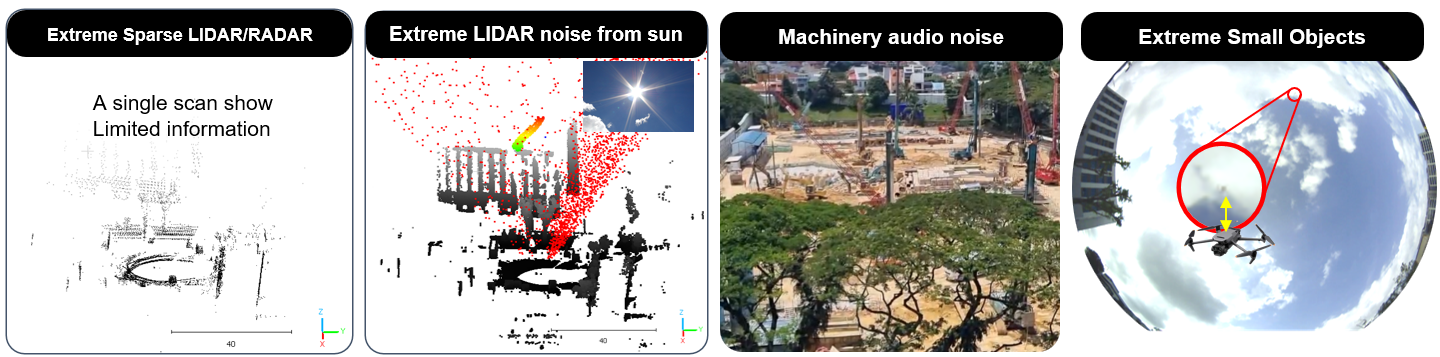}
\caption{
\textbf{Examples of real-world sensing challenges in the MMAUD dataset.} 
(\textbf{Left two}) LiDAR scans suffer from cloudless sunlight interference, introducing significant point cloud noise that reduces spatial reliability. 
(\textbf{Third}) Audio channels are affected by continuous heavy machinery and construction noise, degrading the signal-to-noise ratio in microphone arrays. 
(\textbf{Right}) RGB frames exhibit visual extremes, such as small UAV appearances under wide-FOV distortion, specular glare, and low visual contrast against bright skies. 
These scenarios collectively reflect the need for robust, multimodal perception under challenging real-world conditions.
}
\label{datasetnoises}
\end{figure*}

The dataset also incorporates a range of real-world sensing challenges, illustrated in Fig.~\ref{datasetnoises}, including:

\begin{itemize}
    \item \textbf{LiDAR Noise:} Strong sunlight introduces dynamic interference patterns in LiDAR scans, reducing point cloud reliability.
    \item \textbf{Audio Noise:} Persistent background noise from air conditioning units and nearby construction activity affects acoustic signal integrity.
    \item \textbf{Visual Challenges:} RGB imagery includes extreme cases of small object appearance, specular glare, strong backlighting, and nighttime darkness.
\end{itemize}


\subsection{Baseline Selection}

\begin{table*}[t]
\centering
\caption{For the daytime with idea lighting, the comparison of performance on real-world data in meters.}
\label{tab:realworld_best}
\renewcommand{\arraystretch}{1.15}
\resizebox{\textwidth}{!}{
\begin{tabular}{lccccccccccccc}
\toprule
\textbf{Methods} & \textbf{Year} & \textbf{Label} & \textbf{Category} & \textbf{Modal} &
\multicolumn{4}{c}{\textbf{Current Pose Estimation}} &
\multicolumn{4}{c}{\textbf{Future Pose Prediction}} \\
\cmidrule(lr){4-5}\cmidrule(lr){6-9} \cmidrule(lr){10-13}
& & & (Training) & (Inference) &
$D_x \downarrow$ & $D_y \downarrow$ & $D_z \downarrow$ & $E_{0s} \downarrow$ &
$E_{1s} \downarrow$ & $E_{3s} \downarrow$ & $E_{5s} \downarrow$ \\

\midrule
VisualNet \cite{wang2022visualnet}  & 2022 & TLT & Supervised & Image-Only 
& 1.27 & 1.31 & 1.54 & 2.42 & 4.22 & 6.16 & 8.56 \\
DarkNet \cite{bochkovskiy2020yolov4} & 2020 & TLT & Supervised & Image-Only 
& 1.38 & 1.71 & 2.00 & 2.63 & 4.61 & 7.29 & 9.05 \\
Yolov8 \cite{varghese2024yolov8}     & 2024 & TLT & Supervised & Image-Only 
& 0.83 & 0.98 & 1.18 & 1.62 & 3.73 & 5.34 & 7.71 \\
Yolov10 \cite{wang2024yolov10}       & 2024 & TLT & Supervised & Image-Only 
& \underline{0.76} & 0.95 & \underline{0.75} & \underline{1.44} & 3.10 & 4.94 & 6.79 \\
\midrule

AudioNet \cite{yang2023av}             & 2023 & TLT & Supervised & Audio-Only 
& 1.99 & 2.18 & 2.32 & 3.76 & 6.01 & 10.86 & 16.41 \\
VorasNet \cite{vora2023dronechase}     & 2023 & TLT & Supervised & Audio-Only 
& 1.91 & 2.07 & 2.07 & 3.04 & 6.97 & 11.28 & 14.18 \\
AAUTE \cite{lei2025audio}              & 2025 & TLT & Self-Supervised & Audio-Only 
& 1.74 & 1.82 & 1.78 & 2.64 & 3.78 & 8.66 & 10.79 \\
\midrule
ASDNet  \cite{tao2021someone}          & 2021 & TLT & Supervised & Audio+Image 
& 1.16 & 1.10 & 1.22 & 1.79 & 3.57 & 4.71 & 7.96 \\
AV-PED \cite{yang2023av}               & 2023 & TLT & Self-Supervised & Audio+Image  
& 1.16 & \underline{0.92} & 1.21 & 1.93 & 3.49 & 4.86 & 6.99 \\
AV-FDTI \cite{yang2024av}              & 2024 & TLT & Supervised & Audio+Image  
& 1.07 & \textbf{0.84} & 1.08 & 1.68 & 2.78 & \underline{4.21} & 6.29 \\
\midrule
\multirow{2}{*}{\textbf{Ours}}  & -- & TLT & Self-Supervised & Image-Only 
& \textbf{0.59} & 0.97 & \textbf{0.57} & \textbf{1.27} & \textbf{1.71} & \textbf{1.97} & \textbf{3.85} \\
                                & -- & Proposed & Self-Supervised & Image-Only 
& 1.09 & 1.36 & 1.28 & 1.95 & \underline{2.77} & \underline{4.21} & \underline{6.20} \\
\bottomrule

\end{tabular}
}

\vspace{5pt}
\begin{flushleft}
\footnotesize{RMSE: Error between predicted and actual values. The smaller, the better the estimation.Best results are \textbf{boldened}, and second-best results are \underline{underlined}.} \\
\footnotesize{
\textbf{Proposed:} The pseudo-label noise in our proposed self-supervised framework is estimated to be approximately \(\sim 5\) pixels (\(\sim 27\,\text{cm}\) error at 30 meters) when using commercial off-the-shelf LiDARs. Despite this moderate error, our method remains fully \textbf{label-free and deployable} in real-world UAV scenarios, and is evaluated on the challenging MMAUD V1 and V2 sequences—\textbf{more difficult than those used in prior works}.  
For reference, Terrestrial Laser Tracking (TLT) systems offer sub-centimeter accuracy (typically \(\sim 5\,\text{mm}\)), but are costly and impractical for field deployment.
}
\end{flushleft}

\end{table*}

To ensure a comprehensive evaluation, we selected a diverse
set of baselines, including vision-only, audio-only, and audiovisual fusion pipelines, trained using either supervised or
self-supervised approaches. Our baseline selection prioritized
publicly available code, task relevance, and scalability to 3D
trajectory prediction. We excluded methods from our comparison that exhibit one or more of the following limitations:

\noindent \textbf{Closed-source}: Not released until submission \cite{zheng2024keypoint}, \cite{ding2023drone},

\noindent \textbf{Over-parameterized}: Need extensive param tuning \cite{yu2023unified}, \cite{makansi2020multimodal},

\noindent \textbf{Limited to 2D}: lacking scalability to 3D \cite{lan2018robust}, \cite{li2023adaptive}, \cite{siyuan2024learning}.

For baseline models originally limited to 2D detection, we incorporated an additional 3D estimation head to enable full 3D position inference. For those already capable of 3D estimation, we integrated a shared linear extrapolation module \cite{talbot2024continuous} to support future trajectory prediction. Linear extrapolation was selected for its robustness and minimal parameter tuning, in contrast to more complex alternatives such as B-splines \cite{ding2019efficient}
MINCO \cite{wang2022geometrically}
or Gaussian Processes \cite{lilge2025incorporating}
, which often suffer from overfitting and inconsistent performance across UAV sequences. All baseline and proposed methods are linked for transparency and reproducibility. The evaluated baselines include:

\noindent \textbf{VisualNet}\footnote{VisualNet: \url{https://github.com/doubleblindsubmit/UAV_Comparison}}: VisualNet \cite{wang2022visualnet} is an end-to-end framework built on a CNN backbone for single-image 2D position inference.
We integrated a 3D position regressor and trained in a fully supervised manner. Additional linear extrapolation components
are integrated for future position prediction.

\noindent \textbf{Darknet}\footnote{DarkNet: \url{https://github.com/pjreddie/darknet}}: DarkNet \cite{bochkovskiy2020yolov4} uses CSPDarkNet53, an image-only
network for 2D drone detection. An additional regressor is integrated to estimate the 3D position, while linear extrapolation
components are added for future position prediction.

\noindent \textbf{Yolov8}\footnote{Yolov8: \url{https://github.com/ultralytics/ultralytics}}: YoloV8 \cite{varghese2024yolov8}, a 2D object detection framework, is
employed as the backbone for feature extraction. A subsequent
3D position estimation head is integrated, followed by linear
extrapolation to predict the trajectory.

\noindent \textbf{Yolov10}\footnote{Yolov10: \url{https://github.com/THU-MIG/yolov10}}: YoloV10 \cite{wang2024yolov10}, a lightweight version of the Yolo series for 2D object detection, is similarly used as the backbone.
The extracted features are passed to a 3D position estimation
head, and future trajectory prediction is performed using linear
extrapolation.

\noindent \textbf{AudioNet}\footnote{AudioNet: \url{https://github.com/doubleblindsubmit/UAV_Comparison}}: AudioNet \cite{yang2023av} uses only audio input from AVPED [26], with the audio features directly regressing to the
UAV’s location without fusion. We enhance this by adding a
linear extrapolation component for future position prediction.

\noindent \textbf{VorasNet}\footnote{VorasNet: \url{https://github.com/doubleblindsubmit/UAV_Comparison}}: VorasNet is \cite{vora2023dronechase} a cross-modal network for 3D
drone detection that uses audio-only inference to determine
the current drone position. Due to its relevance to our specific objectives and being closed-source, we had to fully
re-implemented the model. Additional linear extrapolation
components are integrated for future position prediction.

\noindent \textbf{AAUTE}\footnote{AAUTE: \url{https://github.com/AllenLei666/AAUTE}}: AAUTE \cite{lei2025audio} is a self-supervised network with
LiDAR pseudo-labeling to infer 3D UAV position with audioonly signals. Additional linear extrapolation components are
integrated for future position prediction.

\noindent \textbf{ASDNet}\footnote{ASDNe: \url{https://github.com/TaoRuijie/TalkNet-ASD}}: ASDNet \cite{tao2021someone} is a supervised audio-visual active
speaker detection model. We take the audio-visual feature
extraction part as backbone and a regressor for 3D position
estimation. Additional linear extrapolation components are
integrated for future position prediction.

\noindent \textbf{AV-PED}\footnote{AV-PED: \url{https://github.com/yizhuoyang/AV-PedAware}}: AV-PED \cite{yang2023av} is a self-supervised dynamic pedestrian detection network with audio-visual inputs. We replace
our backbone with AV-PED’s and add a regressor to estimate
UAV’s 3D position, and linear extrapolation for future trajectory prediction.

\subsection{Evaluation Metrics}

\begin{table*}[t]
\centering
\caption{For the nighttime environment with low lighting, the comparison of performance on MMAUD in meters.}
\label{tab:mmaud_night}
\renewcommand{\arraystretch}{1.15}
\resizebox{\textwidth}{!}{
\begin{tabular}{lcccccccccccc}
\toprule
\textbf{Methods} & \textbf{Light condition} & \textbf{Category} & \textbf{Modal} &
\multicolumn{4}{c}{\textbf{Current Pose Estimation}} &
\multicolumn{4}{c}{\textbf{Future Pose Prediction}} \\
\cmidrule(lr){3-4}\cmidrule(lr){5-8} \cmidrule(lr){9-12}
& & (Training) & (Inference) & 
$D_x \downarrow$ & $D_y \downarrow$ & $D_z \downarrow$ & $E_{0s} \downarrow$ &
$E_{1s} \downarrow$ & $E_{3s} \downarrow$ & $E_{5s} \downarrow$ \\
\midrule
VisualNet \cite{wang2022visualnet}  & Night & Supervised     & Image-Only   
& 2.28 & 3.05 & 3.30 & 4.75  & 7.40 & 18.80 & 26.55 \\
DarkNet \cite{bochkovskiy2020yolov4} & Night & Supervised     & Image-Only   
& 2.30 & 3.08 & 3.45 & 4.90  & 7.55 & 19.00 & 27.30 \\
Yolov8 \cite{varghese2024yolov8}     & Night & Supervised     & Image-Only   
& 2.10 & 2.95 & 3.15 & 4.45  & 6.85 & 16.90 & 23.75 \\
Yolov10 \cite{wang2024yolov10}       & Night & Supervised     & Image-Only   
& 1.98 & 2.75 & 2.95 & 4.05  & 6.20 & 15.10 & 22.05 \\
\midrule

AudioNet \cite{yang2023av}       & Night & Supervised     & Audio-Only   
& 1.99 & 2.18 & 2.32 & 3.76 & 6.01 & 10.86 & 16.41 \\
VorasNet \cite{vora2023dronechase} & Night & Supervised     & Audio-Only   
& 1.91 & 2.07 & 2.07 & 3.04 & 6.97 & 11.28 & \underline{14.18} \\
AAUTE \cite{lei2025audio}          & Night & Self-Supervised & Audio-Only   
& \textbf{1.74} & \textbf{1.82} & \textbf{1.78} & \textbf{2.64} & \textbf{3.78} & \underline{8.66} & \textbf{10.79} \\
\midrule

ASDNet \cite{tao2021someone}        & Night & Supervised     & Audio+Image  
& 2.18 & 2.50 & 2.65 & 4.10 & 6.60 & 17.50 & 24.00 \\
AV-PED \cite{yang2023av}            & Night & Self-Supervised & Audio+Image  
& 1.95 & 2.35 & 2.40 & 3.10 & 5.20 & 13.60 & 18.65 \\
AV-FDTI \cite{yang2024av}           & Night & Supervised     & Audio+Image  
& \underline{1.88} & \underline{2.05} & \underline{2.10} & \underline{2.85} & 4.60 & 11.05 & 16.75 \\

\midrule

\textbf{Ours}                       & Night & Self-Supervised & 
Image-Only  & 1.89 & 2.30 & 2.13 & 3.25  & \underline{4.28} & \textbf{7.34} & 15.04 \\
\bottomrule
\end{tabular}
}

\vspace{5pt}
\begin{flushleft}
\footnotesize{Performance comparison under low-light nighttime conditions on the MMAUD dataset.
We evaluate both current pose estimation and future trajectory prediction across different methods and modalities.
Despite relying solely on self-supervised learning and image-only input, our method achieves strong performance in future trajectory forecasting, surpassing supervised audio-visual baselines.} \\
\end{flushleft}

\end{table*}

\begin{figure}[!t]
\centering
\includegraphics[width=3.4in]{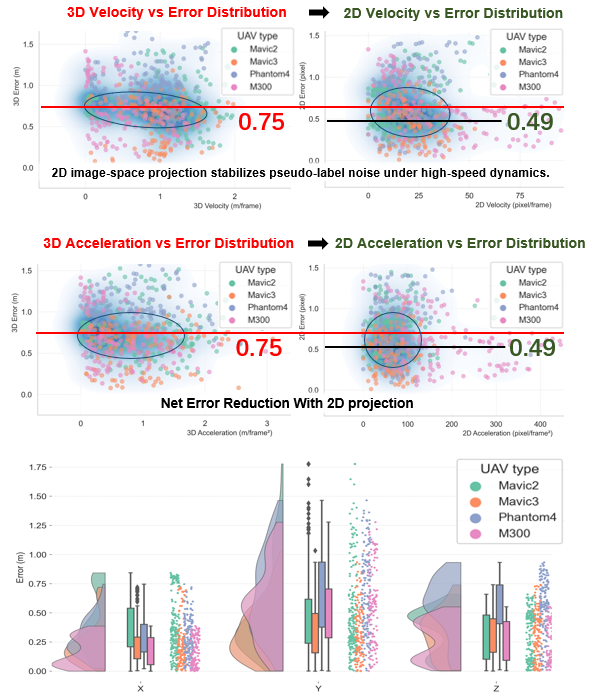}
\caption{
\textbf{Impact of Motion Dynamics on Pseudo-Label Error and the Effect of 2D Projection.}  
We analyze how pseudo-label error correlates with UAV motion dynamics across velocity and acceleration domains. In both cases, 2D image-space projection significantly reduces the mean error (from 0.75\,m to 0.49\,m), especially under high-speed conditions (highlighted ellipses). The bottom plot further shows axis-wise error distributions across four UAV types, confirming the robustness and generalization of our approach under diverse trajectories.
}
\label{Dynamic_Distribution}
\end{figure}

We evaluate our algorithm using the Root Mean Square Error (RMSE) between the predicted and ground-truth UAV positions across multiple time horizons. This metric directly quantifies the spatial accuracy of state estimation and future prediction. In particular, we report RMSE along each axis ($D_x$, $D_y$, $D_z$) as well as the aggregated Euclidean error ($E_{0s}$, $E_{1s}$, $E_{3s}$, $E_{5s}$), following standard UAV evaluation protocols.

To further understand model performance under different UAV dynamics, we analyze prediction error as a function of UAV velocity and acceleration. As shown in Fig.~\ref{Dynamic_Distribution}, we plot the correlation between velocity/acceleration magnitudes and error magnitudes in both 2D and 3D settings. Each point represents a UAV frame, color-coded by platform.

\subsection{Result and Discussion}

\begin{table*}[t]
\captionsetup{justification=raggedright,singlelinecheck=false}
\caption{Ablation study on three core components: C1 (Temporal-KNN), C2 (3D-2D Alignment), and C3 (Using Event Cues). Results are evaluated on future prediction RMSE at 3s and 5s.}
\label{tab:ablation}
\renewcommand{\arraystretch}{1.15}
\setlength{\tabcolsep}{5pt}
\centering
\begin{tabular}{lccccccccc}
\toprule
\textbf{Ablation } & \textbf{C1} & \textbf{C2} & \textbf{C3} & \textbf{Model} & \textbf{$E_{0s}$ ↓} & \textbf{$E_{1s}$ ↓} & \textbf{$E_{3s}$ ↓} & \textbf{$E_{5s}$ ↓} & \textbf{Validation Contribution} \\
\midrule
\textbf{Ours}           & \ding{51} & \ding{51} & \ding{51} & Vision-Mamba & \textbf{1.95} & \textbf{2.77} & \textbf{4.21}  & \textbf{6.20} &  \textbf{Full Method} \\
\hline
w/o Vision-Mamba        & \ding{51} & \ding{51} & \ding{51} & Transformer & 2.01 & 2.86 & 4.67  & 6.42 &  Validates \textbf{Vision-Mamba} \\

\hline
w/o Temporal-KNN                      & \ding{53}     & \ding{51}     & \ding{51}     & Vision-Mamba & 2.14 & 3.29 & 4.61           & 7.45     & Validates \textbf{Temporal-KNN}      \\
\hline
w/o 3D-2D Alignment                     & \ding{51} & \ding{53}     & \ding{51} & Vision-Mamba & 3.97 & 5.16 & 6.40           & 9.74     & Validates \textbf{3D-2D Alignment}     \\
\hline
w/o Event Cues                         & \ding{51} & \ding{51} & \ding{53}     & Vision-Mamba & 2.15 & 2.93 & 4.27           & 6.75      & Validates \textbf{Effect of Event Cues}     \\

\bottomrule

\end{tabular}

\vspace{3pt}
\footnotesize{
Bold indicates best performance. \ding{51} means component is enabled, \ding{53} means removed. Each ablation validates the corresponding contribution C1, C2, or C3.
}
\end{table*}





We conduct comprehensive evaluations across four UAV platforms: DJI Mavic 2, Mavic 3, Phantom 4, and M300. The results, summarized in Table~\ref{tab:realworld_best}, cover both current pose estimation and multi-step trajectory forecasting under ideal daytime lighting conditions.

Our method consistently achieves competitive or superior performance compared to both supervised and self-supervised baselines, while relying solely on image inputs and pseudo-labels for training. We achieve \textbf{6.20m} RMSE at 5 seconds future prediction, outperforming YOLOv8 (7.71m), AudioNet (16.41m), and AV-FDTI (6.29m), despite requiring no manual 3D annotations. These results highlight the strength of our label-free learning pipeline, which combines 3D temporal structure with motion-aligned visual supervision (C1–C2).

Fig.~\ref{Results} visualizes the effect of incorporating dynamics-aware prediction constraints. The top row shows error growth across future time steps; the middle rows compare trajectory realism with and without constraint modeling. Without motion-aware supervision, forecasts tend to drift or overshoot. In contrast, our constrained predictions are more physically plausible and temporally consistent. \textbf{The bottom row quantifies this improvement in mean and variance across UAV types}.

In addition, Fig.~\ref{Dynamic_Distribution} analyzes how prediction error varies with motion velocity and acceleration, supporting the robustness of our self-supervised labels under dynamic UAV behaviors. Despite using pseudo-labels derived from unsynchronized and noisy LiDAR data, our alignment mechanism (C2) and event-informed learning (C3) enable the model to maintain low error even under high-speed motion.

Importantly, these results are achieved using a \textbf{scalable, monocular RGB-based pipeline} with no reliance on LiDAR at test time. This makes the method highly deployable for real-time counter-UAV scenarios where size, cost, and power constraints preclude the use of 3D sensors.

Overall, the framework offers a practical and label-efficient solution for long-horizon UAV trajectory forecasting, with direct implications for airspace defense, kinetic interception, and regulatory drone monitoring in urban or infrastructure-sensitive environments.

\begin{figure}[!t]
\centering
\includegraphics[width=3.4in]{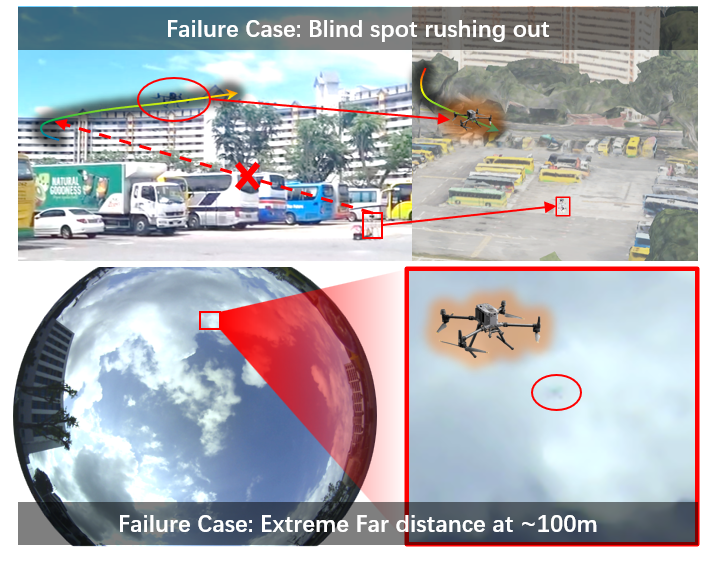}
\caption{Visual illustration of two edge-case scenarios that challenge our framework: (top) a UAV rapidly emerging from an occluded region near a vehicle roof, and (bottom) a distant UAV appearing as a sub-pixel target in the fisheye view. These conditions reduce motion visibility and degrade pseudo-label reliability.}
\label{failurecases}
\end{figure}

\section{Limitations and Future Directions}
\subsection{Failure Cases and Edge-Condition Limitations}
While our method demonstrates strong overall performance, we observe failure cases in specific edge conditions as shown in Fig. \ref{failurecases}. Notably, prediction errors increase when the UAV flies at very low altitudes—such as near the ground, behind trees, or close to the top of large vehicles like buses—placing it within blind spots of the detection rig. Failures also occur when the UAV flies at extreme distances (e.g., beyond 100 meters), where it appears as only a few pixels in the image. In both cases, visual and motion cues become unreliable, leading to degraded pseudo-label quality and downstream prediction accuracy due to limited observable dynamics.

\subsection{Future Directions} 
\noindent \textbf{Lack of Real Fisheye Events:} Our current framework simulates event cues by computing intensity changes from high-FPS RGB images (1280$\times$720), as commercially available wide-FOV, high-resolution event cameras—such as the recently released Prophesee GenX720—were not accessible during data collection. While this simulation captures coarse motion dynamics and provides useful supervisory signals, it does not reflect the true characteristics of event-based sensing. As shown in Fig.~\ref{simulated_data}, real event cameras offer precise asynchronous response, high temporal resolution, and exceptional dynamic range—qualities that simulated events cannot replicate. As a result, our current system represents a conservative lower-bound performance with the added noise from simulated events. The simulated cues are significantly noisier and less temporally accurate, especially under fast motion or extreme lighting conditions. In practice, replacing simulated input with real event streams is expected to substantially improve motion fidelity and prediction accuracy. Future work will focus on collecting real wide-FOV event data under conditions consistent with MMAUD to enhance modality realism, reduce label noise, and improve generalization.

\noindent \textbf{Limited Environment Test:} Evaluations are performed only on the MMAUD dataset, which is limited to rooftop and carpark scenes. Although it includes diverse UAV platforms and sensor configurations, the lack of coverage across open-field terrain, forest or rural areas limits generalization. {A key extension would be to benchmark on multi-domain datasets or conduct real-world field tests in more diverse and challenging terrains.}

\noindent \textbf{Heuristic Parameter Sensitivity:}  
Several components in our framework rely on empirically chosen parameters, such as the temporal gradient threshold $\tau$ in KNN filtering, the weighting factors in motion alignment loss, and the selection of temporal windows for event cue simulation. While these settings yield robust results across the MMAUD dataset, we do not conduct systematic sensitivity analyses, and their adaptability to new sensing setups or environments remains unverified. Future work may incorporate differentiable hyperparameter optimization or meta-learning~\cite{messikommer2024reinforcement} techniques for adaptive tuning in unseen domains.

\section{Conclusion}

This paper presents a label-free, self-supervised framework for 3D UAV trajectory forecasting using only monocular RGB and simulated event cues. The proposed pipeline introduces three key innovations: (i) a Temporal-KNN algorithm for extracting motion-consistent trajectories from raw, asynchronous LiDAR without manual annotation; (ii) a cross-modal projection alignment mechanism that refines pseudo-labels using ORB and event-based visual motion cues; and (iii) a Vision-Mamba-based predictor trained via stage-wise self-supervision to recover long-horizon 3D UAV trajectories under diverse flight conditions.

Extensive experiments on the MMAUD dataset demonstrate that our method achieves state-of-the-art forecasting performance, outperforming both supervised image, audio, and fusion-based baselines—without relying on any manual 3D labels. The system generalizes across UAV types, lighting conditions, and flight dynamics, while remaining fully deployable in real-time without LiDAR or radar at inference. We also identify and analyze key failure cases under low-altitude occlusion and extreme-range tracking, which inform directions for future robustness enhancements.

By addressing challenges of label scarcity, sensor asynchrony, and limited depth perception, this work provides a scalable, cost-effective solution for proactive UAV interception and airspace monitoring. Future extensions will explore broader environment deployment, integration of real fisheye event sensors, and adaptive parameter tuning via reinforcement learning or differentiable optimization.

\bibliographystyle{IEEEtran}
\bibliography{mybib}






\end{document}